\documentclass[runningheads]{llncs}
\usepackage[T1]{fontenc}
\usepackage{graphicx}
\usepackage{hyperref}
\usepackage{xcolor} 
\newcommand{\now}{\textit{present point}} 
\newcommand{\nows}{\textit{present points}} 
\newcommand{\seccase}{complete-set} 
\newcommand{\shortname}{LongForMAD} %
\newcommand{\longname}{Longitudinal Forecasting Model for Alzheimer's Disease} %
\newcommand{\changecolor}{black} 
\newcommand{\revision}[1]{\textcolor{\changecolor}{#1}}

\begin{document}
\title{\revision{Assessing the significance of longitudinal data in Alzheimer's Disease forecasting}}
\titlerunning{\shortname}


\author{Batuhan K. Karaman\inst{1,2} \and
Mert R. Sabuncu\inst{1,2}}
\institute{Cornell University and Cornell Tech, New York, NY 10044, USA \and
Weill Cornell Medicine, New York, NY 10021, USA}

\maketitle              
\begin{abstract}
In this study, we \revision{employ a} transformer encoder model \revision{to characterize the significance of longitudinal patient data for} forecasting the progression of Alzheimer’s Disease (AD). 
Our model, {\longname} ({\shortname}), harnesses the comprehensive temporal information embedded in sequences of patient visits that incorporate multimodal data, providing a deeper understanding of disease progression than can be drawn from single-visit data alone.
We present an empirical analysis across two patient groups—Cognitively Normal (CN) and Mild Cognitive Impairment (MCI)—over a span of five follow-up years. 
Our findings reveal that models incorporating more extended patient histories can outperform those relying solely on present information, suggesting a deeper historical context is critical in enhancing predictive accuracy for future AD progression. 
Our results support the incorporation of longitudinal data in clinical settings to enhance the early detection and monitoring of AD. Our code is available at \href{https://github.com/batuhankmkaraman/LongForMAD}{https://github.com/batuhankmkaraman/{\shortname}}.

\keywords{Alzheimer's Forecasting \and Longitudinal Data \and Transformer Neural Networks}
\end{abstract}
\section{Introduction \revision{and Related Literature}}
Alzheimer's Disease (AD) poses a significant challenge to global healthcare systems, affecting millions of individuals and their families. 
Early and accurate prediction of AD progression is crucial for effective management and treatment planning. 
The two key progression events to forecast in Alzheimer's are the conversion from the cognitively normal (CN) state to the mild cognitive impairment (MCI) state, and from the MCI state to the AD state.
Recent advancements in machine learning and deep learning have opened new avenues for predicting disease progression, leveraging vast amounts of medical data. 
Longitudinal studies have been instrumental in identifying risk factors and progression markers of various disease types \cite{lee_2023_enhancing}.

\revision{
In previous studies, various methods have been employed to leverage longitudinal data for AD.
Longitudinal cortical thickness changes from multiple time points in patients' history are used for classifying CN vs AD and sMCI (stable MCI) vs pMCI (progressive MCI) at the time of the most recent visit in \cite{li_2012_discriminant} with a support vector machine (SVM). 
\cite{zhang_2012_predicting} uses longitudinal clinical data and biomarkers from multiple time points to classify sMCI vs pMCI using a multi-kernel SVM. 
In more recent years, \cite{cui_2019_rnnbased} employs an RNN-based model for CN vs AD and sMCI vs pMCI classification using MRIs collected at multiple time points. 
\cite{jarrett_2020_dynamic} employs a convolutional architecture for survival prediction, using longitudinal clinical data and biomarkers collected from multiple time points in patients' history to forecast non-AD vs AD outcomes in a 5-year future time horizon.}
\revision{
Transformer encoders \cite{vaswani_2017_attention}, have shown promise in medical data analysis by capturing the temporal dynamics of disease progression \cite{Wang2023.06.28.23291994,shen_2023_leveraging}. 
Their capability to process variable-length and non-uniform input sequences makes them well-suited for analyzing heterogeneous longitudinal historical data commonly found in healthcare records. 
In the context of Alzheimer's, \cite{hu_2023_vggtswinformer} employs a transformer-based model for sMCI vs pMCI classification using 3D MRIs from the patient's current visit and one prior visit. 
Similarly, \cite{chen_2023_longformer} utilizes a transformer-based model to combine image embeddings extracted from MRIs for CN vs AD classification at the time of the last MRI. 
However, relying solely on the patient's current visit and one prior visit is inadequate for comprehensively understanding the importance of longitudinal data.}
\revision{
The aforementioned studies either do not predict future clinical state or focus solely on sMCI vs pMCI for the MCI group or non-AD vs AD discrimination, without distinguishing between CN and MCI groups. 
The sMCI and pMCI classifications are based on an arbitrary follow-up year. 
While some studies use all years in a 5-year follow-up \cite{sajjadfouladvand_2023_machine}, they do not thoroughly investigate the importance of longitudinal patient data across different prediction time horizons. 
Few studies explore the progression of CN baseline patients \cite{chen_2017_progression}, and none have fully examined the impact of longitudinal patient history, such as history duration and data collection frequency.
}

\revision{
In this work, we quantify the significance of longitudinal patient data in AD forecasting for both CN-baseline and MCI-baseline patient groups, across each year of a 5-year follow-up time horizon, and for various longitudinal data modalities. 
We utilize an extensive history of patients' visits, going back up to three years, and assess the impact of history on a model's predictions in terms of both history duration and data collection frequency. 
To facilitate this analysis, we present a {\longname} ({\shortname}), a transformer encoder-based model capable of making predictions under diverse data modality and patient history availability scenarios for both CN-baseline and MCI-baseline patients and for any future time point.
Our analysis highlights the critical importance of longitudinal data in clinical environments, illustrating its pivotal role in enhancing the early detection of AD, especially in the CN cohort.
}




\section{Materials and Methods}

\subsection{Dataset}
All participants used in this work are from the Alzheimer's Disease Neuroimaging Initiative (ADNI) database \cite{mueller_2005_ways}. 

Participants are chosen based on their clinical diagnosis status. 
Specifically, we include individuals who have not been diagnosed with clinical AD at their baseline visit.
Additionally, these participants must have undergone at least one follow-up diagnostic assessment after this baseline visit. 
We exclude CN baseline subjects who converted to AD because they are very few (n=9). 
In addition, we exclude CN baseline participants who converted to MCI before reverting to CN (n=41), and MCI baseline participants who were diagnosed as CN in a later follow-up year (n=284) since these subjects might have been diagnosed incorrectly at some point. 
Therefore, we focus solely on CN baseline subjects who progress to MCI and MCI baseline subjects who advance to AD, reflecting the irreversible progression of Alzheimer's disease.
After the exclusions, we are left with 1404 participants.
Table~\ref{baselinedemog} lists summary statistics for the participants.

\begin{table}[hbtp]
    \centering
    {\caption{Summary statistics of the participants at baseline. Mean $\pm$ standard deviations are listed. APOE4 row represents the number of alleles. APOE4, Apolipoprotein E4; CDR, Clinical Dementia Rating; MMSE, Mini Mental State Examination. }\label{baselinedemog}}
    {\begin{tabular}{lll}
    \hline
    \bfseries ~& \bfseries CN at & \bfseries MCI at\\
    \bfseries ~& \bfseries baseline & \bfseries baseline\\
    \bfseries ~& \bfseries $(n=615)$ & \bfseries $(n=789)$\\
    \hline
    Female/Male & 335 / 280 & 324 / 465\\
    Age $(yr)$ & 73.19 $\pm$ 6.18 & 73.46 $\pm$ 7.39\\
    Education $(yr)$ & 16.51 $\pm$ 2.57 &  15.93 $\pm$ 2.81\\
    APOE4 $(0/1/2)$ & 430/169/14 & 371/313/98\\
    CDR & 0.04 $\pm$ 0.13 & 1.55 $\pm$ 0.89\\
    MMSE & 29.11 $\pm$ 1.11 & 27.52 $\pm$ 1.82\\
    \hline
    \end{tabular}}
\end{table}

The data, typical of many longitudinal studies, includes missing follow-up visits, irregular timings, and subject dropouts before study completion. 
Table \ref{labeldistribution} details the number of subjects in each diagnostic group for annual follow-ups. 
In all analyses, including Table \ref{labeldistribution}, subjects who progressed to a later stage (CN to MCI, or MCI to AD) before dropping out are assumed to remain at the advanced stage. 
Stable subjects not present in a specific follow-up year are excluded from training and testing for that year.

\begin{table}[hbtp]
  \centering
  {\caption{The number of available subjects in each diagnostic group for annual follow-up visits. The follow-up diagnoses are not actually exactly 12 months apart. They have been rounded to the nearest time horizon in years. DX, Diagnosis.} \label{labeldistribution}}
  \begin{tabular}{l|l|lllllllllllllll}
  \hline
  \bfseries Patient & \bfseries Follow-up & \multicolumn{14}{c}{\bfseries Follow-up year} \\
  \bfseries Group & \bfseries DX & \bfseries 1 & \bfseries 2 & \bfseries 3 & \bfseries 4 & \bfseries 5 & \bfseries 6 & \bfseries 7 & \bfseries 8 & \bfseries 9 & \bfseries 10 & \bfseries 11 & \bfseries 12 & \bfseries 13 & \bfseries 14 & \bfseries 15 \\
  \hline
  CN at & CN & 427 & 527 & 181 & 230 & 123 & 173 & 97 & 72 & 33 & 17 & 23 & 11 & 19 & 2 & 4 \\\cline{2-17}
  baseline & MCI & 14 & 32 & 41 & 49 & 54 & 65 & 79 & 83 & 85 & 87 & 87 & 88 & 88 & 88 & 88 \\
  \hline
  MCI at & MCI & 674 & 431 & 317 & 202 & 127 & 93 & 88 & 57 & 30 & 20 & 6 & 8 & 3 & - & - \\\cline{2-17}
  baseline & AD & 110 & 218 & 261 & 286 & 292 & 305 & 313 & 322 & 324 & 326 & 327 & 327 & 327 & 327 & 327 \\
  \hline
  \end{tabular}
\end{table}

\subsection{Input features}
\label{sec:inputfeatures}
We use the clinical data and biomarkers as our input features. 
Clinical data includes subject demographics, genotype (the number of APOE4 alleles, specifically), cognitive test scores, and diagnosis (CN or MCI). 
The complete list of demographic features and cognitive tests can be found in ~\cite{adnimanual}.
The biomarkers are Cerebrospinal Fluid (CSF) measurements and Magnetic Resonance Imaging (MRI) volume measurements \cite{jack_2015_magnetic} (computed using the FreeSurfer software \cite{Fischl01012004}). 
In the ADNI study, subjects undergo multiple cognitive assessments and MRI scans across visits, providing a detailed longitudinal medical history. 
The demographic and genotype variables remain static, and CSF is not collected at every visit. 
ADNI is divided into four phases (ADNI-1, ADNI-GO, ADNI-2, and ADNI-3), each with its own data acquisition protocol. 
There is considerable heterogeneity in the follow-up data collection, including irregular visit intervals and frequent missing visits. 
The extent of data missingness across various modalities for the 1404 participants in our study is detailed in Table~\ref{missingness_entire}.



\begin{table}[hbtp]
    \centering
    
    {\caption{The degree of missingness (\%) in different data modalities for two patient groups throughout the entire enrollment period. COGN, Cognitive Tests; MRI, Magnetic Resonance Imaging; CSF, Cerebrospinal Fluid.} \label{missingness_entire}}
    {\begin{tabular}{lll}
    \hline
    \bfseries Data & \bfseries CN at & \bfseries MCI at\\
    \bfseries type & \bfseries baseline & \bfseries baseline\\
    \hline
    COGN & 45.28 & 43.42 \\
    MRI & 58.22 & 48.00 \\
    CSF & 85.54 & 83.97 \\
    \hline
    \end{tabular}}
\end{table}

Following \cite{karaman_2022_machine}, we begin the input preprocessing by recording the binary missingness mask for the feature set. 
Each participant has their own binary missingness mask indicating what variable was observed or not for that particular individual at a specific visit. 
Then, we perform mode/mean substitution for missing categorical/numerical features, respectively, following \cite{ADNIImputation}. 
The categorical variables except the diagnosis are one-hot encoded, and numerical variables are z-score normalized in the last step of feature processing. 
To prevent any information leakage, both mode/mean substitution and z-score normalization statistics are collected from training data and used for training, validation and test data. 
Concatenating the input features, and the binary missingness mask yields a feature vector of length 113.

\subsection{Model} 
\begin{figure*}[htb]
  {\includegraphics[width=\linewidth]{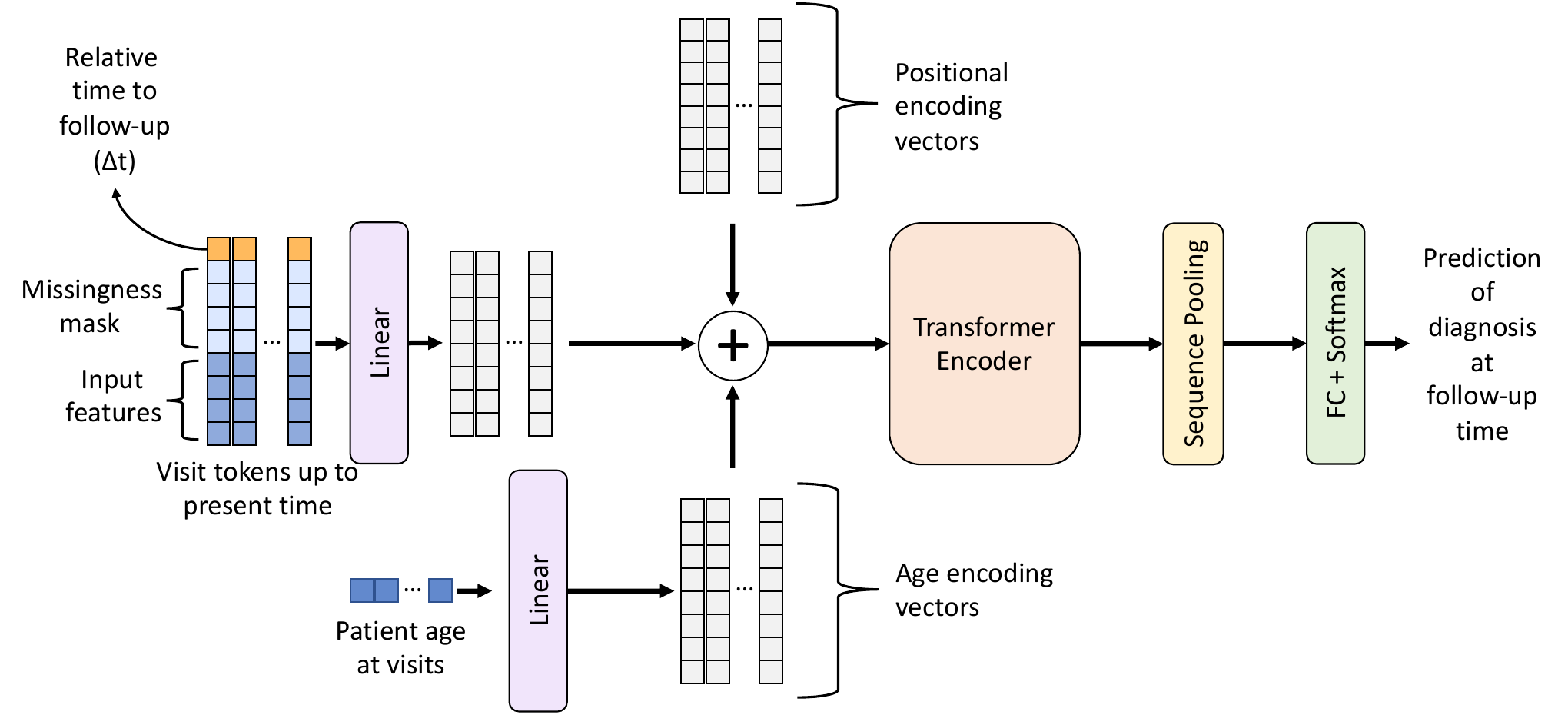}}
  {\caption{Schematic representation of the {\longname} ({\shortname}) architecture, illustrating the workflow from input to the final prediction output.}\label{modelfig}}
\end{figure*}

We focus on predicting an individual's future diagnostic status, categorizing it into one of three classes: CN, MCI, or AD. 
This prediction is based on data obtained from preceding annual medical visits. 
The chronological nature of these visits forms the basis of our sequential input, with the goal of generating a corresponding classification prediction. 
To achieve this, we implement a transformer-based sequence-to-class neural network, {\shortname}, wherein each medical visit is represented as a distinct input token. 
The detailed architecture of {\shortname} is illustrated in Fig.~\ref{modelfig}.

Our model enhances each visit token by appending the prediction time horizon ($\Delta$t in Figure~\ref{modelfig}), defined as the interval between the visit's data collection and the intended future prediction time point. 
Following \cite{karaman_2022_machine}, this increases the input token length to 114. 
This adaptation allows our model to make predictions over various future time horizons without the need to train distinct networks for each horizon. 
Instead, a singular, comprehensive network is trained across the entire dataset, covering all labels from subsequent follow-up years.
In terms of network architecture, we employ a transformer encoder layer, a sequence pooling layer, and a fully connected classifier. 
Notably, our network integrates a standard learnable position encoding layer (initialized with 0 vectors) and a parameterized linear age encoding layer, as proposed by \cite{li_2020_behrt}.

Following the addition of positional and age encodings to the input visit tokens, which are projected to the same latent space by a linear layer, the transformer encoder processes the input utilizing self-attention.
This process enables the efficient identification and extraction of pertinent features from each token by analyzing the patient's AD longitudinal data comprehensively. 
The sequence pooling layer, which applies average pooling along the token axis, then consolidates the information gathered from the entire longitudinal data of the patient. 
The aggregated feature set is subsequently passed through the fully connected classifier with softmax activation, which computes a diagnosis probability distribution, which represents the probabilities of CN, MCI, and AD diagnoses at the predicted future time point.
Our network incorporates element-wise rectified linear units (ReLUs) as nonlinear activation functions between layers.

\subsection{Training} 
\label{sec:training}
Transformers, despite their effectiveness in various tasks, are notably data intensive and prone to overfitting due to their extensive use of self-attention mechanisms, which require substantial data to learn effectively. 
We have designed a training strategy incorporating several techniques to mitigate overfitting and enhance model performance.

To begin, we expand our training and validation datasets. 
We consider every visit with a diagnosis of CN or MCI within the training data as a present-time reference point, denoted as {\now}. 
Visits with a confirmed AD diagnosis are not designated as {\now} in our process, as projecting their future is beyond the purview of this study.
For each visit designated as {\now}, we compile the patient's historical data, which may extend to a maximum of three years prior. 
Consequently, the longest sequence of longitudinal data used for training encompasses four visits: the visit occurring three years before {\now}, the visit occurring two years before {\now}, the visit one year before {\now}, and the visit at the point of {\now}.
Subsequently, we record all subsequent follow-up diagnoses occurring in the five years succeeding {\now}. 
By constructing these 9-year progression trajectories, we effectively triple the volume of training and validation samples, diversifying the dataset. 
To prevent information leakage, we ensure that all 9-year progression trajectories originating from the same subject are exclusively allocated to one of the splits: either training, validation, or test set. 

We leverage the inherent capability of transformer encoders to process input sequences of variable lengths, utilizing this feature to perform data augmentation on longitudinal history records. 
We perform the augmentation by omitting randomly selected past visits from the training data in each training epoch. 
Such a technique prompts the model to generalize its learning across the full spectrum of a patient's longitudinal data, rather than narrowly adapting to particular visits, thereby reducing the potential for overfitting. 

\revision{
Table~\ref{labeldistribution} reveals two imbalances: the changing class label distribution over time and the decreasing availability of clinical labels due to dropouts. 
The fluctuation in the number of stable CN baseline individuals across different years can be linked to varying monitoring policies in the ADNI study phases. 
To address these imbalances, we employ a loss re-weighting scheme during training, similar to \cite{karaman_2022_machine}, which assigns higher penalties to misclassifications in minority classes by using sample-level weights. 
We categorize participants into four groups for each follow-up year (CN baseline non-converters, CN baseline converters, MCI baseline non-converters, and MCI baseline converters) and adjust the weights to ensure balanced representation across all years. 
Without this mitigation, the model might prioritize predicting follow-up years and patient groups with the highest number of samples, neglecting the minorities.
} 
The weights for each sample point are calculated based on the expanded datasets. 

We calculate a single training loss using the sample points generated by our augmentation strategy.
However, for validation, we evaluate the validation loss across every possible scenario regarding the presence or absence of a visit in a history year, which amounts to $2^4-1=15$ different scenarios. 
These scenarios include cases such as whether the visit at {\now} occurred, whether there was a visit 1 year before {\now}, and so forth. 
To generate these scenarios, we simply exclude the relevant visits from the history of the validation subjects.
We then compute the average of these loss values and use this average as the criterion for early stopping. 



\subsection{Evaluation}
\label{sec:evaluation}
In evaluation, our primary goal is to assess the influence of the presence of patient longitudinal history on the predictions made by our model.
It is important to highlight that our transformer-based network is designed to handle cases where a patient's longitudinal history is not fully complete, meaning it does not necessarily include all past visits before {\now}. 
This enables the model to generate predictions even with partial or no historical data.
\revision{
We perform the analysis across various longitudinal data modality cases present in the input. Our model is capable of making predictions with missing features in the visits, as discussed in Section~\ref{sec:inputfeatures}. 
To create those ``cases", we synthetically replace unwanted features with mean/mode values derived from the training set and flip the corresponding entries in the missingness mask.}
We calculate our performance scores in two patient groups (CN at {\now} and MCI at {\now}) and every follow-up year separately. 
Therefore, each score analysis we conduct is bifurcated, one corresponding to the CN-to-MCI conversion task and the other to the MCI-to-AD conversion task. 

ADNI, like many real-world longitudinal studies, is subject to biases in subject recruitment and follow-up. 
A notable issue is ``temporal bias" \cite{yuan_2021_temporal}, arising from the non-uniform distribution of visits across disease stages. 
\revision{Employing an inference strategy that mitigates temporal bias is crucial when evaluating model performance.}
To compute the performance scores for a particular patient group, follow-up year, \revision{longitudinal data modality case}, and longitudinal history scenario, the process begins by randomly selecting a single {\now} for each test subject. 
This selection is made by choosing from all available instances of {\nows}, with each selected instance ensuring that the subject belongs to the specific patient group of interest (CN at {\now} or MCI at {\now}).
Furthermore, each chosen instance should be linked with a diagnosis in the designated follow-up year. 
It is important to highlight that within a pseudo test set, each subject contributes a single follow-up diagnosis to be predicted from a certain historical instance.
Thus, all sample points in a pseudo test set are independent. 
\revision{Then, we obtain predictions by utilizing all relevant input features and longitudinal history information implied by the longitudinal data modality case and history scenario.} 
We record the \revision{performance scores} and repeat this operation for multiple random pseudo test sets.
In the final stage, we calculate the average of the \revision{performance score values} obtained from those pseudo test sets. 
This average represents the final comprehensive score for that specific patient group, follow-up year, and longitudinal history scenario pair.
The purpose of repeating the operation for multiple pseudo test sets is to broaden the range of disease progression trajectories we use in our evaluation.
\revision{By doing so, we aim to mitigate the potential effects of temporal bias on our results.
We note that since our dataset expansion strategy utilizes every possible visit as {\now} during training, as discussed in Section~\ref{sec:training}, it not only expands the dataset size but also addresses temporal bias.}

\section{Experiments}

\subsection{Experimental Details}
We employed a randomized, diagnosis-stratified approach to divide the data into training and testing sets with an 80-20 ratio. 
This division process was replicated 200 times, with the results presented herein being the average outcomes of these iterations. 
Within each split, we conducted a 5-fold cross-validation on the training set, and the validation loss was used to determine early stopping for model training. 
For each fold, five models were trained with different random seeds. 
The final predictions for each test scenario were obtained by averaging the outputs of 25 models, combining 5 cross-validation folds and 5 unique initializations.
We used Adam optimizer \cite{kingma_2017_adam} for training. 
To tune the architecture of {\shortname}, we employ a grid search strategy across each of the  train/test splits. 
The optimal architecture for each test set is determined based on the hyperparameter values that achieve the highest performance on a corresponding validation set. 
The details of fixed hyperparameters, along with the optimization grid used for the adjustable hyperparameters, are documented on our github repository. 
For the calculation of performance scores, we utilize 100 random pseudo test sets.
Due to the unbalanced follow-up diagnoses in Table~\ref{labeldistribution}, we primarily use the area under the receiver operating characteristic curve (AUROC) to evaluate our model. 
Our analyses show nearly identical ROC curves for both one-versus-one and one-versus-rest approaches, indicating a low likelihood of CN subjects being predicted as AD and MCI subjects as CN. 
Therefore, we report the one-versus-rest analysis results, considering MCI as the positive class for CN-to-MCI conversions and AD for MCI-to-AD conversions.

\subsection{Results}
A comparison of {\shortname}'s performance with other existing longitudinal models in the literature for MCI-to-AD conversion in the third follow-up year is presented in Table~\ref{benchmarks}. 
It is important to note that our comparison is limited to this specific patient group and follow-up year because the models against which we can benchmark our performance are primarily focused on this particular scenario. 
We also emphasize that this comparison is made primarily to demonstrate that our modeling approach is sufficiently robust, justifying the analysis of its predictions.

\begin{table*}[hbtp]
  \centering
  {\caption{Comparison of {\shortname}'s performance with other existing models in the literature. MRI; magnetic resonance imaging, COGN; cognitive assessments.}\label{benchmarks}}
  {\begin{tabular}{l|l|l|l}
  \hline
  ~ & Longitudinal & Sequential & 3-year \\
  ~ & data & network & MCI-to-AD \\
  Model & modality & architecture & AUROC \\
  \hline
  \cite{cui_2019_rnnbased} & MRI & RNN & 73.03\\
  \hline
  \cite{li_2019_a} & MRI & - & 74.60\\
  \hline
  \cite{altay_2020_preclinical} & MRI & - & 77.97\\
  \hline
  \cite{altay_2020_preclinical} & MRI & - & 79.26\\
  \hline
  \cite{hu_2023_vggtswinformer} & MRI & Transformer & 81.53\\
  \hline
  {\shortname} (Ours) & MRI & Transformer & 78.65\\
  \hline
  {\shortname} (Ours) & MRI $+$  & Transformer & 92.03\\
  ~ &  COGN & ~ & ~\\
  \hline
  \end{tabular}}
\end{table*}

We assess the impact of patient longitudinal history availability in two ways: 
We compare the performance of our model when patient history information is available starting from years -3, -2, -1, and 0 to investigate the impact of longitudinal history duration, year 0 being the {\now}. 
Then, by keeping the history duration constant at 2 years (i.e., using history starting from year -2), we analyze the effect of longitudinal history data collection frequency.
We compare the model's performance when data is collected annually (involving years -2, -1, and 0) versus biennially (involving years -2 and 0). 
We note that in the analysis of history duration, we maintain a constant data collection frequency of 1 visit per year.

\revision{
As detailed in Section~\ref{sec:inputfeatures}, the modalities that present a longitudinal history are cognitive test scores and MRI biomarkers.
We assess the importance of patient longitudinal history across two time-varying (i.e., longitudinal) data modality ``cases.''
In the first case, referred as ``MRI-only", we synthetically induce missingness in cognitive test scores for all patients, leaving MRI biomarkers as the sole time-varying modality, to assess the significance of historical MRI biomarkers. 
Then, for a comprehensive assessment of the impact of longitudinal history, we incorporate all available input features in the second scenario, referred as ``{\seccase}" case. 
}

\subsubsection{Impact of longitudinal history duration}
\label{sec:results_dur} 

\begin{figure}[htb]
  \includegraphics[width=1\linewidth]{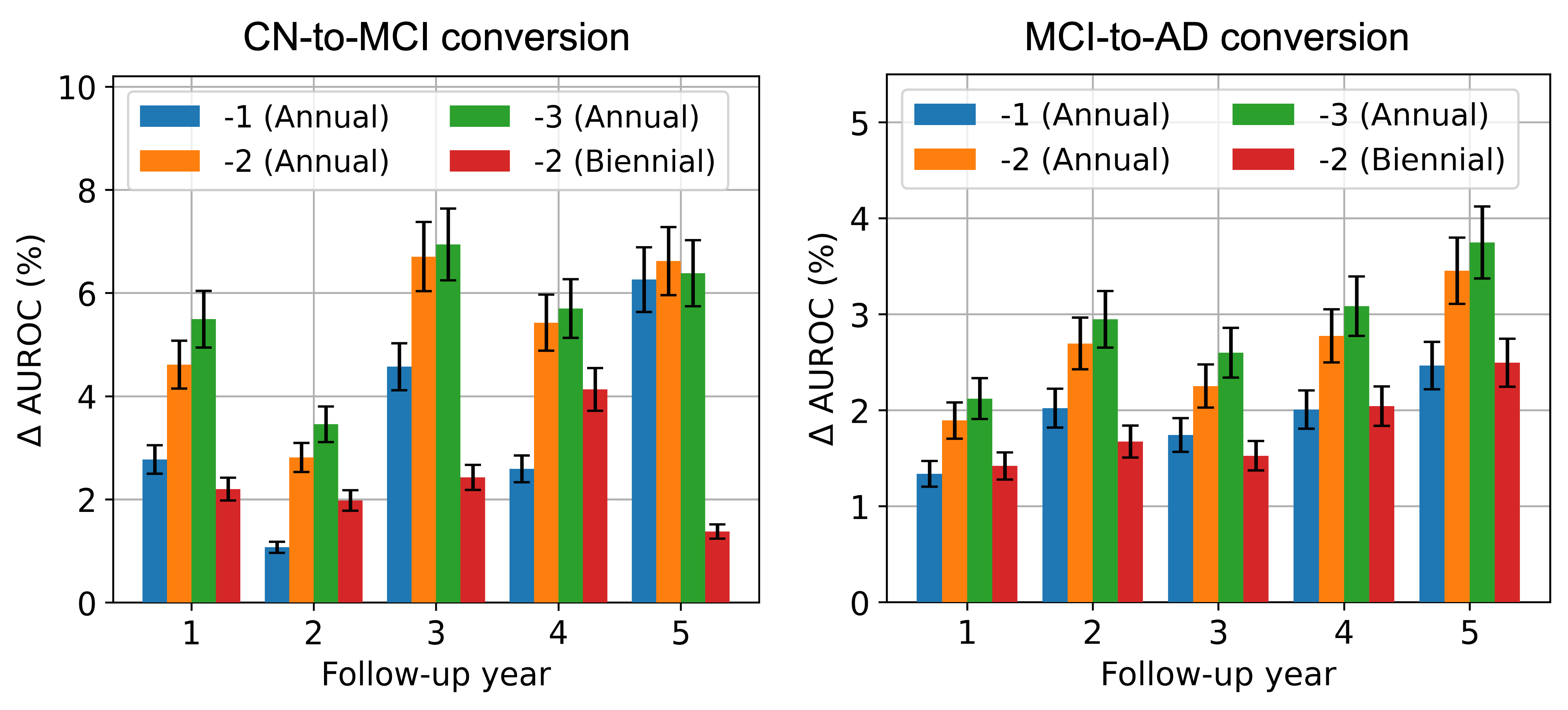}%
  {\caption{\revision{$\Delta$AUROC scores obtained with the addition of longitudinal MRI data. Legend format is ``longitudinal history start year (data collection frequency)", year 0 being the {\now}. Error bars indicate the standard error.}}\label{fig:figuremri}}
\end{figure}

\revision{
\subsubsection{MRI-only case}
Figure~\ref{fig:figuremri} shows the change in AUCROC ($\Delta$AUROC) scores obtained with the addition of historical MRI data against the scenario where no longitudinal history is available.
In Figure~\ref{fig:figuremri}, incorporating longitudinal MRIs consistently improves prediction performance for CN-to-MCI conversion across all follow-up years. 
The largest improvement is observed in the third year of follow-up. 
We note that the benefits of a 3-year annual history appear to diminish slightly compared to a 2-year annual history.
In fact, incorporating the visit data from year -3 results in a decrease in performance in follow-up year 5.
This suggests that including patient history over longer durations might lead to overfitting, likely due to the inclusion of time points that offer no predictive value.}
\revision{
For MCI-to-AD conversion, the addition of prior MRIs again leads to an increase in AUROC scores. 
Extending the history duration yields a diminishing return in performance improvement, similar to the CN-to-MCI conversion; however, a decrease in performance is not observed. 
Moreover, unlike CN-to-MCI conversion, the value of historical MRIs for MCI-to-AD conversion appears to increase almost monotonically as the prediction time horizon extends. 
This suggests that the earliest changes leading to AD are  prominently reflected in longitudinal MRIs.}


\subsubsection{Complete-set case}
\begin{figure}[htb]   
  \includegraphics[width=1\linewidth]{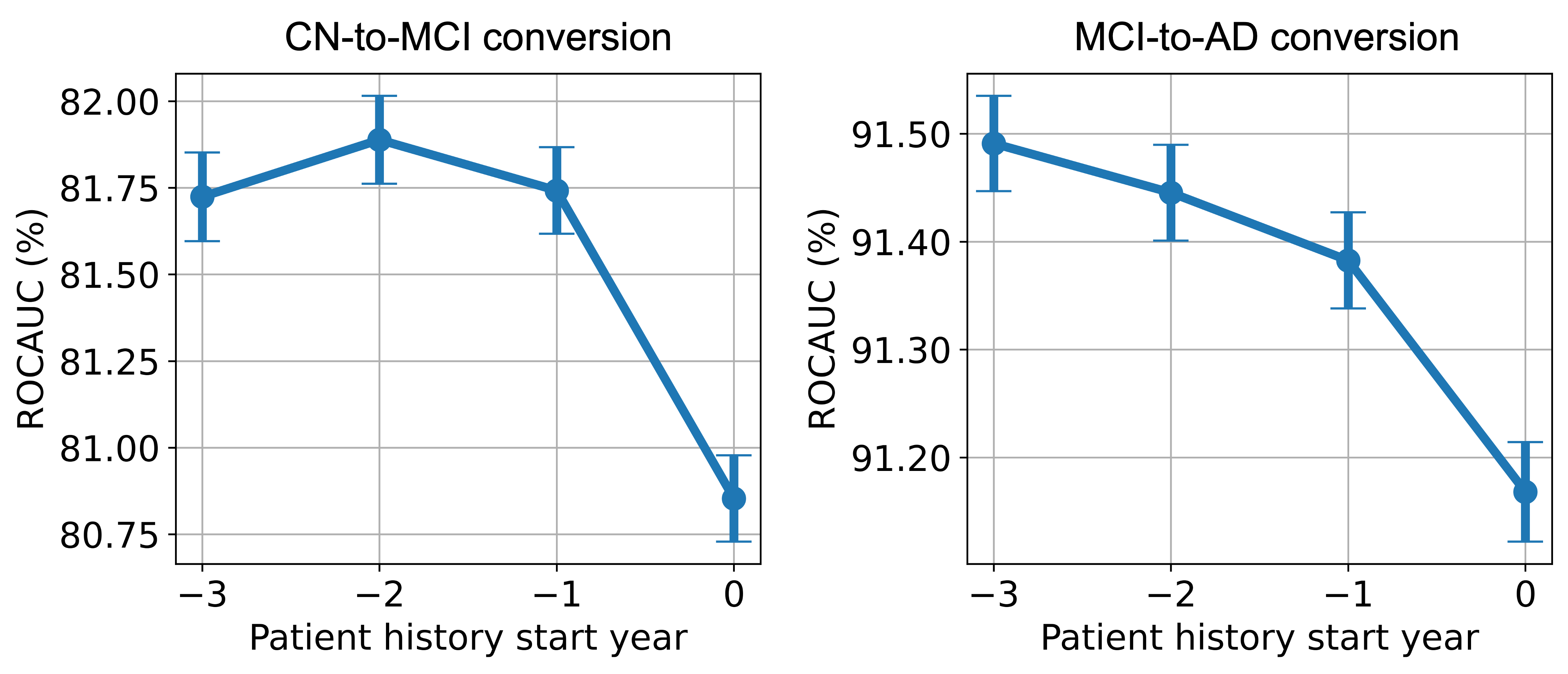}%
  {\caption{AUROC scores for CN to MCI and MCI to AD conversion with up to three years of longitudinal history inclusion for the input feature set containing all data modalities. The x-axis represents the start year of the patients' longitudinal history relative to {\now}, year 0 being the {\now}. AUROC values are averaged across 5 follow-up years. Error bars indicate the standard error.}\label{fig:figuredur}}
\end{figure}

In the {\seccase}, i.e., when both longitudinal MRI and cognitive variable sets are included, we observe similar patterns in the AUROC gains of the different follow-up years. 
Hence, we only show the average AUROC score (obtained as the mean over the five follow-up years) in Figure~\ref{fig:figuredur}. 


In Figure~\ref{fig:figuredur}, we observe that for CN-to-MCI conversion, the inclusion of a full set of data modalities from the patient's history does indeed enhance the model's performance. 
\revision{However, the magnitude of improvement is notably smaller compared to the MRI-only case in Figure~\ref{fig:figuremri}. 
This indicates the cognitive assessments are highly predictive for AD forecasting, and having access to such data at the {\now} alone brings the model's prediction performance close to its potential maximum.}
Compared to the MRI-only scenario, diminishing returns are more evident when historical data is extended from 1 to 3 years. 
The inclusion of data from year -3 results in reduced model performance averaged over all follow-up years, unlike the performance drop which occurred in only the fifth follow-up year in MRI-only case. 
This reduction in AUROC scores with broader history use implies that older data may be less relevant, possibly clouding the more recent, indicative information. 

\revision{The MCI-to-AD conversion panel of Figure \ref{fig:figuredur} shows a significantly smaller increase in performance with the inclusion of history compared to the MRI-only case in Figure~\ref{fig:figuremri}, highlighting the predictive strength of cognitive test scores.}
Consistent with our earlier MRI-only findings, we do not observe a decrease in performance within the 3-year history window.


\subsection{Impact of data collection frequency}
We examine the effect of longitudinal history data collection frequency by maintaining a constant history duration of two years and comparing $\Delta$AUROC scores between scenarios of annual and biennial data collection against the scenario with no patient history.
\revision{
For the MRI-only case depicted in Figure~\ref{fig:figuremri}, it's evident that annual data collection offers a significantly greater improvement in predictive performance compared to biennial collection for both patient groups. 
Additionally, it is worth noting that the gain in performance with annual data collection is smaller in MCI-to-AD conversion when compared to CN-to-MCI conversion within the MRI-only case.}

\begin{figure}[htb]
  \includegraphics[width=\linewidth]{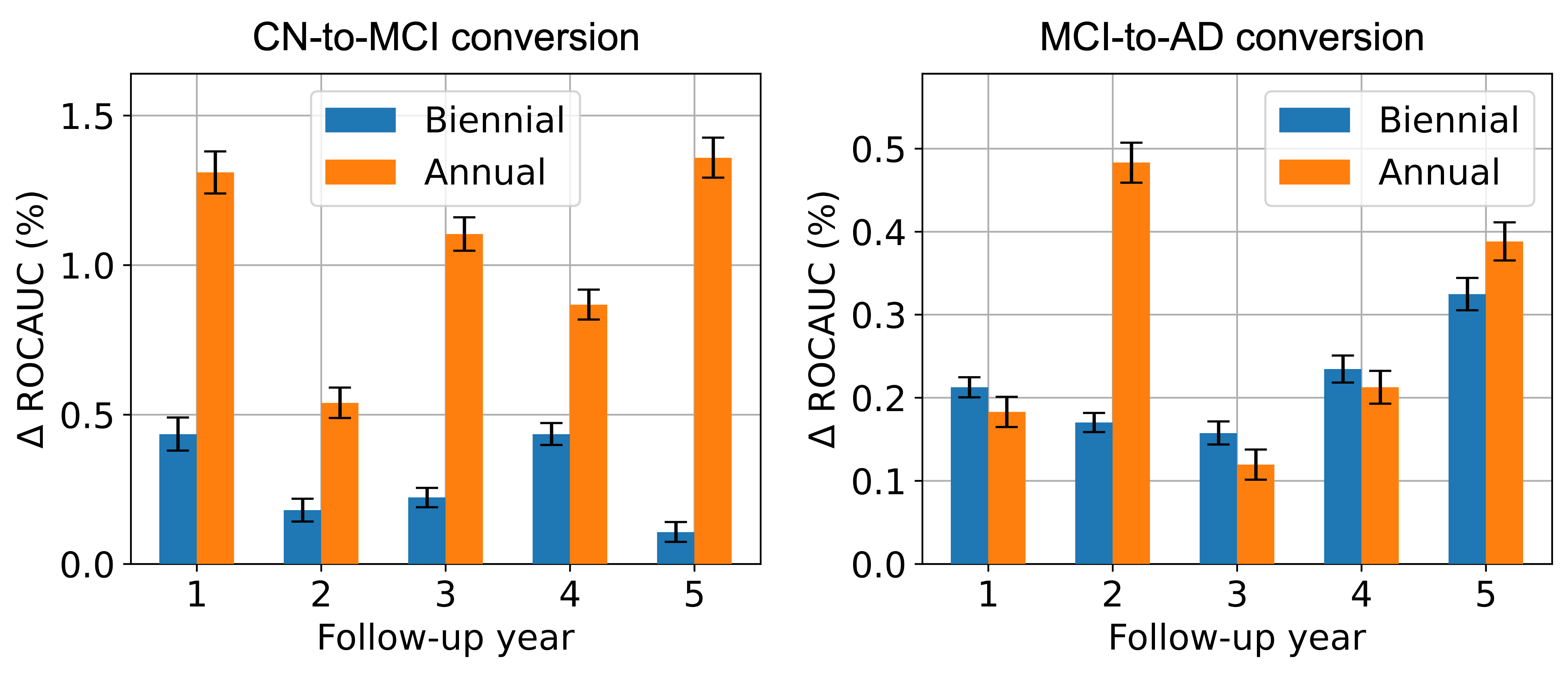} 
  \caption{$\Delta$~AUROC scores obtained with the addition of biennial and annual longitudinal history data to the {\now} data for the input feature set containing all data modalities. The history duration is 2 years. Error bars indicate the standard error across these splits.}
  \label{fig:figurefreq}
\end{figure}

Figure~\ref{fig:figurefreq} illustrates a similar comparison for the {\seccase} case.
For the CN-to-MCI conversion, the trends are nearly identical to those observed in the MRI-only case, albeit with a smaller overall increase in performance. 
For the MCI-to-AD conversion, the differences between annual and biennial data collection frequencies are less stark, yet on average, annual data collection shows a slight advantage, particularly due to a notable improvement in the second follow-up year. 
The results demonstrate that both CN and MCI patient groups benefit from higher frequencies of historical data collection, with the CN group showing a more pronounced and consistent improvement in AUROC in both data modality cases. 
This again highlights that more sophisticated modeling, such as including longitudinal history, offers a larger boost for the harder CN-to-MCI conversion prediction task, as stated in Section~\ref{sec:results_dur}.


\section{Conclusion}
\revision{In this work, we presented a comprehensive analysis of AD forecasting with the use of patient's longitudinal clinical data and biomarker histories.}  
The results we obtained with {\shortname} reveal a significant benefit from incorporating more extensive longitudinal history, particularly for those with CN, who demonstrated a more substantial improvement in predictive performance. \revision{The observed trends from multiple data modalities encourage further investigation into personalized monitoring and intervention strategies based on imaging methods.
In the future, incorporating image encoders into our model would be valuable for extracting more nuanced features from MRIs. 
Additionally, visualizing the changes in the localization of these encoders with respect to the changes in longitudinal history presence could provide valuable insights into disease progression. 
}



%
%
%
\bibliographystyle{splncs04}
\bibliography{mybibliography}
%




\end{document}


\title{Supplementary Material for Assessing the significance of longitudinal data in Alzheimer's Disease forecasting}
\titlerunning{Supplementary Material for \shortname}


\author{Batuhan K. Karaman\inst{1,2} \and
Mert R. Sabuncu\inst{1,2}}
\institute{Cornell University and Cornell Tech, New York, NY 10044, USA \and
Weill Cornell Medicine, New York, NY 10021, USA}
\maketitle              

\subsection{Additional information about input features}
\label{app:inputfeatures}
Demographic features in ADNI are age, gender, number of years of education completed, ethnicity, race, marital status.
Cognitive tests are clinical dementia rating, or CDR; Activities of Daily Living, or FAQ; Everyday Cognition, or ECog, Mini-Mental State Exam, or MMSE; Alzheimer’s Disease Assessment Scale, or ADAS-Cog; Montreal Cognitive Assessment, or MoCA; Rey Auditory Verbal Learning Test Trials 1–6; Logical Memory Delayed Recall; Trail Making Test Part B; Digit Symbol Substitution, Digit and Trails B versions of Preclinical Alzheimer’s Cognitive Composite score.

\subsection{Additional Experimental Details}
\label{app:experimentaldetails}
We used the Adam optimizer with a learning rate set at \(5e-4\) and incorporated an L2 penalty loss on the weights and biases with a regularization weight of \(1e-4\). 
The training utilized a batch size of 32. 
We used a global dropout probability of \(0.5\).
For longitudinal history sequence augmentation through visit dropping, we set the probability of augmentation application at 0.8, meaning there is an 80\% chance that any given patient history sequence will undergo augmentation. 
When augmentation is applied, each visit within the sequence has a 50\% probability of being omitted.

To optimize the {\shortname} architecture, we conducted a \(2 \times 2 \times 2 \times 2\) grid search over key parameters: the hidden dimension, number of attention heads, number of layers in the transformer encoder, and the depth of the final fully-connected classifier network. 
The search parameters for the transformer encoder were divided into grids of \(\{128, 256\}\) for the hidden dimension, \(\{1, 2\}\) for the number of heads, and \(\{1, 2\}\) for the number of layers. 
For the fully-connected classifier network, we evaluated two configurations: \(\{ [128, 3], [3] \}\). 
Here, \([128, 3]\) indices a structure with a linear layer of width 128 followed by a ReLU activation and a subsequent linear layer of width 3 post sequence pooling. 
The \([3]\) configuration points a singular linear layer of width 3 following sequence pooling.




\subsection{Individual AUROC scores of the history scenarios presented in Figure~2}
\label{app:mri}
Please refer to Table~\ref{tab:rocauc_mri}.

\begin{table}[hbtp]

  {\caption{AUROC scores obtained with {\shortname} under various longitudinal history scenarios for the input modality case with MRI as the only longitudinal modality. History scenario format is ``history start year (data collection frequency)", year 0 being the {\now}. AUROC format is ``mean $\pm$ standard error".}\label{tab:rocauc_mri}}  
  {\begin{tabular}{l|l|l|l|l|l|l|l|l|l|l|l}
  \hline
  Long. & \multicolumn{1}{l|}{History} & \multicolumn{5}{l|}{\bfseries CN-to-MCI conversion} & \multicolumn{5}{l}{\bfseries MCI-to-AD conversion }\\\cline{3-12}
  mod. & scenario & 1-year & 2-year & 3-year & 4-year & 5-year & 1-year & 2-year & 3-year & 4-year & 5-year\\
  \hline
  MRI & 0       & 51.10 &        54.96 &        57.85 &        64.41 &        62.92 &         72.43 &         74.05 &         76.40 &         80.87 &         80.12 \\
  ~   & -       & (0.73) &      (0.56) &      (0.52) &      (0.47) &      (0.48) &       (0.27) &       (0.26) &       (0.27) &       (0.28) &       (0.31) \\
  \hline
  MRI & -1      & 53.87 &        56.04 &        62.42 &        67.01 &        69.18 &         73.76 &         76.07 &         78.14 &         82.88 &         82.59 \\
  ~   & (Annu.) & (0.75) &      (0.58) &      (0.52) &      (0.48) &      (0.48) &       (0.27) &       (0.26) &       (0.27) &       (0.27) &        (0.3) \\
  \hline
  MRI & -2      & 55.71 &        57.78 &        64.55 &        69.84 &        69.54 &         74.32 &         76.74 &         78.65 &         83.64 &         83.58 \\
  ~   & (Annu.) & (0.79) &      (0.59) &      (0.54) &       (0.5) &      (0.51) &       (0.26) &       (0.26) &       (0.26) &       (0.27) &       (0.29) \\
  \hline
  MRI & -3 & 56.59 &        58.42 &        64.79 &        70.11 &        69.31 &         74.55 &         76.99 &         79.00 &         83.95 &         83.87 \\
  ~   & (Annu.) & (0.80) &      (0.59) &      (0.55) &       (0.5) &      (0.51) &       (0.26) &       (0.26) &       (0.26) &       (0.26) &       (0.29) \\
  \hline
  MRI & -2      & 53.30 &        56.94 &        60.27 &        68.55 &        64.30 &         73.85 &         75.72 &         77.92 &         82.91 &         82.62 \\
  ~   & (Bien.) & (0.77) &      (0.58) &      (0.54) &      (0.49) &      (0.52) &       (0.27) &       (0.25) &       (0.26) &       (0.27) &       (0.29) \\
  \hline
  \end{tabular}}
\end{table}

\subsection{Detailed metrics for the complete-set case}
\label{app:comb}
Refer to Table~\ref{tab:rocauc_comb} for AUROC, Table~\ref{tab:bacc_comb} for balanced accuracy, Table~\ref{tab:sens_comb} for sensitivity,
Table~\ref{tab:spec_comb} for specificity, Table~\ref{tab:prauc_comb} for area under the precision-recall curve (AUPR), and Table~\ref{tab:ece_comb} for expected calibration error (ECE).

\begin{table}[hbtp]

  {\caption{AUROC scores obtained with {\shortname} under various patient history scenarios for the input modality case with MRI and cognitive tests (MRI$+$COGN) as the longitudinal modalities (complete-set case). History scenario format is ``history start year (data collection frequency)", year 0 being the {\now}. AUROC format is ``mean $\pm$ standard error".}\label{tab:rocauc_comb}}  
  {\begin{tabular}{l|l|l|l|l|l|l|l|l|l|l|l}
  \hline
  Long. & \multicolumn{1}{l|}{History} & \multicolumn{5}{l|}{\bfseries CN-to-MCI conversion} & \multicolumn{5}{l}{\bfseries MCI-to-AD conversion }\\\cline{3-12}
  mod. & scenario & 1-year & 2-year & 3-year & 4-year & 5-year & 1-year & 2-year & 3-year & 4-year & 5-year\\
  \hline
  MRI $+$ & 0       & 80.86 &        81.96 &        80.06 &        80.21 &        81.18 &         86.48 &         89.19 &         91.92 &         94.14 &         94.11 \\
  COGN   & -        & (0.43) &      (0.34) &       (0.30) &      (0.31) &      (0.32) &       (0.14) &       (0.12) &       (0.12) &       (0.12) &       (0.13) \\
  \hline
  MRI $+$ & -1      & 81.97 &        82.44 &        81.01 &        80.75 &        82.54 &         86.63 &         89.63 &         92.02 &         94.24 &         94.40 \\
  COGN   & (Annu.)  & (0.42) &      (0.34) &       (0.30) &      (0.31) &      (0.32) &       (0.13) &       (0.12) &       (0.12) &       (0.12) &       (0.12) \\
  \hline
  MRI $+$ & -2      & 82.17 &        82.50 &        81.16 &        81.08 &        82.53 &         86.67 &         89.67 &         92.03 &         94.35 &         94.50 \\
  COGN   & (Annu.)  & (0.42) &      (0.34) &      (0.31) &      (0.32) &      (0.33) &       (0.13) &       (0.12) &       (0.12) &       (0.11) &       (0.12) \\
  \hline
  MRI $+$ & -3      & 82.15 &        82.44 &        80.99 &        80.76 &        82.27 &         86.70 &         89.71 &         92.10 &         94.40 &         94.55 \\
  COGN   & (Annu.)  & (0.42) &      (0.34) &      (0.31) &      (0.32) &      (0.33) &       (0.13) &       (0.12) &       (0.12) &       (0.12) &       (0.12) \\
  \hline
  MRI $+$ & -2      & 81.29 &        82.14 &        80.28 &        80.65 &        81.28 &         86.70 &         89.36 &         92.07 &         94.38 &         94.44 \\
  COGN   & (Bien.)  & (0.43) &      (0.35) &       (0.30) &      (0.32) &      (0.33) &       (0.14) &       (0.12) &       (0.12) &       (0.11) &       (0.12) \\
  \hline
  \end{tabular}}
\end{table}

\begin{table}[hbtp]

  {\caption{Balanced accuracy scores obtained with {\shortname} under various patient history scenarios for the input modality case with MRI and cognitive tests (MRI$+$COGN) as the longitudinal modalities (complete-set case). History scenario format is ``history start year (data collection frequency)", year 0 being the {\now}. Balanced accuracy format is ``mean $\pm$ standard error".}\label{tab:bacc_comb}}  
  {\begin{tabular}{l|l|l|l|l|l|l|l|l|l|l|l}
  \hline
  Long. & \multicolumn{1}{l|}{History} & \multicolumn{5}{l|}{\bfseries CN-to-MCI conversion} & \multicolumn{5}{l}{\bfseries MCI-to-AD conversion }\\\cline{3-12}
  mod. & scenario & 1-year & 2-year & 3-year & 4-year & 5-year & 1-year & 2-year & 3-year & 4-year & 5-year\\
  \hline
  MRI $+$ & 0       & 72.16 &      73.14 &      71.47 &      71.76 &      72.40 &       76.59 &       80.64 &       83.18 &       86.81 &       86.80 \\
  COGN   & -        & (0.54) &    (0.37) &    (0.32) &    (0.31) &    (0.31) &     (0.15) &     (0.15) &     (0.17) &     (0.18) &      (0.2) \\
  \hline
  MRI $+$ & -1      & 72.93 &      73.62 &      71.97 &      72.39 &      73.74 &       76.93 &       80.81 &       83.60 &       87.29 &       87.43 \\
  COGN   & (Annu.)  & (0.54) &    (0.36) &    (0.32) &    (0.31) &    (0.32) &     (0.15) &     (0.15) &     (0.17) &     (0.18) &      (0.2) \\
  \hline
  MRI $+$ & -2      & 72.77 &      73.54 &      72.04 &      72.38 &      73.58 &       77.00 &       80.90 &       83.74 &       87.40 &       87.74 \\
  COGN   & (Annu.)  & (0.53) &    (0.37) &    (0.32) &    (0.31) &    (0.32) &     (0.14) &     (0.15) &     (0.17) &     (0.18) &      (0.2) \\
  \hline
  MRI $+$ & -3      & 72.72 &      73.49 &      71.85 &      71.97 &      73.19 &       77.13 &       80.99 &       83.80 &       87.30 &       87.70 \\
  COGN   & (Annu.)  & (0.53) &    (0.37) &    (0.32) &    (0.31) &    (0.32) &     (0.14) &     (0.15) &     (0.17) &     (0.18) &     (0.19) \\
  \hline
  MRI $+$ & -2      & 72.13 &      73.22 &      71.66 &      71.93 &      72.55 &       76.83 &       80.86 &       83.47 &       87.14 &       87.47 \\
  COGN   & (Bien.)  & (0.53) &    (0.37) &    (0.32) &    (0.31) &    (0.31) &     (0.15) &     (0.15) &     (0.17) &     (0.18) &      (0.2) \\
  \hline
  \end{tabular}}
\end{table}

\begin{table}[hbtp]

  {\caption{Sensitivity scores obtained with {\shortname} under various patient history scenarios for the input modality case with MRI and cognitive tests (MRI$+$COGN) as the longitudinal modalities (complete-set case). We consider MCI and AD as the positive class for CN-to-MCI and MCI-to-AD conversion, respectively. History scenario format is ``history start year (data collection frequency)", year 0 being the {\now}. Sensitivity format is ``mean $\pm$ standard error".}\label{tab:sens_comb}}  
  {\begin{tabular}{l|l|l|l|l|l|l|l|l|l|l|l}
  \hline
  Long. & \multicolumn{1}{l|}{History} & \multicolumn{5}{l|}{\bfseries CN-to-MCI conversion} & \multicolumn{5}{l}{\bfseries MCI-to-AD conversion }\\\cline{3-12}
  mod. & scenario & 1-year & 2-year & 3-year & 4-year & 5-year & 1-year & 2-year & 3-year & 4-year & 5-year\\
  \hline
  MRI $+$ & 0       & 73.38 &      73.27 &      74.03 &      73.86 &      75.58 &       93.95 &       91.73 &       90.61 &       90.28 &       90.06 \\
  COGN   & -        & (1.16) &    (0.82) &    (0.69) &    (0.62) &    (0.59) &     (0.23) &     (0.21) &     (0.21) &     (0.22) &     (0.22) \\
  \hline
  MRI $+$ & -1      & 73.81 &      73.60 &      73.02 &      73.71 &      75.32 &       93.44 &       90.79 &       89.86 &       89.37 &       89.09 \\
  COGN   & (Annu.)  & (1.16) &    (0.82) &    (0.68) &    (0.63) &    (0.59) &     (0.23) &     (0.21) &     (0.21) &     (0.22) &     (0.23) \\
  \hline
  MRI $+$ & -2      & 73.74 &      73.57 &      73.53 &      73.99 &      75.70 &       93.07 &       90.41 &       89.55 &       89.03 &       88.65 \\
  COGN   & (Annu.)  & (1.15) &    (0.82) &    (0.68) &    (0.64) &     (0.6) &     (0.23) &     (0.22) &     (0.22) &     (0.23) &     (0.24) \\
  \hline
  MRI $+$ & -3      & 73.55 &      73.22 &      72.96 &      73.21 &      74.82 &       92.94 &       90.23 &       89.23 &       88.35 &       87.99 \\
  COGN   & (Annu.)  & (1.15) &    (0.82) &    (0.67) &    (0.63) &    (0.59) &     (0.23) &     (0.22) &     (0.22) &     (0.23) &     (0.24) \\
  \hline
  MRI $+$ & -2      & 73.31 &      73.31 &      74.21 &      74.15 &      75.71 &       93.39 &       91.11 &       89.99 &       89.57 &       89.22 \\
  COGN   & (Bien.)  & (1.15) &    (0.82) &    (0.69) &    (0.63) &    (0.59) &     (0.23) &     (0.21) &     (0.22) &     (0.22) &     (0.23) \\
  \hline
  \end{tabular}}
\end{table}

\begin{table}[hbtp]

  {\caption{Specificity scores obtained with {\shortname} under various patient history scenarios for the input modality case with MRI and cognitive tests (MRI$+$COGN) as the longitudinal modalities (complete-set case). We consider MCI and AD as the positive class for CN-to-MCI and MCI-to-AD conversion, respectively. History scenario format is ``history start year (data collection frequency)", year 0 being the {\now}. Specificity format is ``mean $\pm$ standard error".}\label{tab:spec_comb}}  
  {\begin{tabular}{l|l|l|l|l|l|l|l|l|l|l|l}
  \hline
  Long. & \multicolumn{1}{l|}{History} & \multicolumn{5}{l|}{\bfseries CN-to-MCI conversion} & \multicolumn{5}{l}{\bfseries MCI-to-AD conversion }\\\cline{3-12}
  mod. & scenario & 1-year & 2-year & 3-year & 4-year & 5-year & 1-year & 2-year & 3-year & 4-year & 5-year\\
  \hline
  MRI $+$ & 0       & 70.94 &      73.02 &      68.91 &      69.66 &      69.22 &       59.22 &       69.55 &       75.75 &       83.34 &       83.54 \\
  COGN   & -        & (0.4) &    (0.38) &    (0.41) &    (0.47) &    (0.53) &     (0.29) &      (0.3) &     (0.34) &     (0.35) &      (0.4) \\
  \hline
  MRI $+$ & -1      & 72.05 &      73.63 &      70.91 &      71.07 &      72.16 &       60.42 &       70.83 &       77.35 &       85.21 &       85.77 \\
  COGN   & (Annu.)  & (0.41) &    (0.37) &    (0.42) &    (0.46) &    (0.53) &     (0.28) &      (0.3) &     (0.33) &     (0.33) &     (0.38) \\
  \hline
  MRI $+$ & -2      & 71.80 &      73.52 &      70.55 &      70.78 &      71.45 &       60.93 &       71.38 &       77.93 &       85.77 &       86.83 \\
  COGN   & (Annu.)  & (0.41) &    (0.37) &    (0.43) &    (0.46) &    (0.53) &     (0.28) &     (0.31) &     (0.33) &     (0.33) &     (0.37) \\
  \hline
  MRI $+$ & -3      & 71.90 &      73.76 &      70.75 &      70.73 &      71.56 &       61.31 &       71.75 &       78.38 &       86.24 &       87.42 \\
  COGN   & (Annu.)  & (0.41) &    (0.37) &    (0.42) &    (0.46) &    (0.53) &     (0.28) &     (0.31) &     (0.32) &     (0.33) &     (0.36) \\
  \hline
  MRI $+$ & -2      & 70.96 &      73.13 &      69.11 &      69.71 &      69.39 &       60.27 &       70.60 &       76.95 &       84.70 &       85.71 \\
  COGN   & (Bien.)  & (0.4) &    (0.38) &    (0.42) &    (0.48) &    (0.53) &     (0.28) &      (0.3) &     (0.33) &     (0.34) &     (0.38) \\
  \hline
  \end{tabular}}
\end{table}

\begin{table}[hbtp]

  {\caption{AUPR scores obtained with {\shortname} under various patient history scenarios for the input modality case with MRI and cognitive tests (MRI$+$COGN) as the longitudinal modalities (complete-set case). History scenario format is ``history start year (data collection frequency)", year 0 being the {\now}. AUPR format is ``mean $\pm$ standard error".}\label{tab:prauc_comb}}  
  {\begin{tabular}{l|l|l|l|l|l|l|l|l|l|l|l}
  \hline
  Long. & \multicolumn{1}{l|}{History} & \multicolumn{5}{l|}{\bfseries CN-to-MCI conversion} & \multicolumn{5}{l}{\bfseries MCI-to-AD conversion }\\\cline{3-12}
  mod. & scenario & 1-year & 2-year & 3-year & 4-year & 5-year & 1-year & 2-year & 3-year & 4-year & 5-year\\
  \hline
  MRI $+$ & 0       & 98.55 &       97.72 &       94.71 &       92.78 &       90.14 &        62.61 &        82.65 &        91.60 &        96.04 &        97.22 \\
  COGN   & -        & (0.04) &     (0.06) &     (0.11) &     (0.16) &     (0.24) &      (0.32) &      (0.21) &      (0.14) &      (0.08) &      (0.07) \\
  \hline
  MRI $+$ & -1      & 98.68 &       97.84 &       94.99 &       93.09 &       90.85 &        63.20 &        84.14 &        91.78 &        96.19 &        97.40 \\
  COGN   & (Annu.)  & (0.03) &     (0.06) &     (0.11) &     (0.16) &     (0.24) &      (0.31) &      (0.19) &      (0.13) &      (0.08) &      (0.06) \\
  \hline
  MRI $+$ & -2      & 98.73 &       97.86 &       95.06 &       93.19 &       90.92 &        63.56 &        84.22 &        91.89 &        96.33 &        97.48 \\
  COGN   & (Annu.)  & (0.03) &     (0.06) &     (0.11) &     (0.16) &     (0.24) &      (0.31) &      (0.19) &      (0.13) &      (0.08) &      (0.06) \\
  \hline
  MRI $+$ & -3      & 98.72 &       97.85 &       94.98 &       93.03 &       90.75 &        63.66 &        84.26 &        91.98 &        96.38 &        97.52 \\
  COGN   & (Annu.)  & (0.03) &     (0.06) &     (0.11) &     (0.17) &     (0.25) &      (0.31) &      (0.19) &      (0.13) &      (0.07) &      (0.06) \\
  \hline
  MRI $+$ & -2      & 98.64 &       97.78 &       94.79 &       92.97 &       90.30 &        63.50 &        83.01 &        92.00 &        96.36 &        97.45 \\
  COGN   & (Bien.)  & (0.03) &     (0.06) &     (0.11) &     (0.16) &     (0.24) &      (0.32) &      (0.21) &      (0.13) &      (0.07) &      (0.06) \\
  \hline
  \end{tabular}}
\end{table}

\begin{table}[hbtp]

  {\caption{Expected Calibration Error (ECE, \%) scores obtained with {\shortname} under various patient history scenarios for the input modality case with MRI and cognitive tests (MRI$+$COGN) as the longitudinal modalities (complete-set case). History scenario format is ``history start year (data collection frequency)", year 0 being the {\now}. ECE score format is ``mean $\pm$ standard error".}\label{tab:ece_comb}}  
  {\begin{tabular}{l|l|l|l|l|l|l|l|l|l|l|l}
  \hline
  Long. & \multicolumn{1}{l|}{History} & \multicolumn{5}{l|}{\bfseries CN-to-MCI conversion} & \multicolumn{5}{l}{\bfseries MCI-to-AD conversion }\\\cline{3-12}
  mod. & scenario & 1-year & 2-year & 3-year & 4-year & 5-year & 1-year & 2-year & 3-year & 4-year & 5-year\\
  \hline
  MRI $+$ & 0       & 21.29 &      21.18 &      21.51 &      22.67 &      25.73 &       58.24 &       43.66 &       36.49 &       31.50 &       26.50 \\
  COGN   & -        & (0.17) &    (0.19) &     (0.2) &    (0.21) &    (0.23) &     (0.09) &      (0.1) &     (0.13) &     (0.17) &     (0.18) \\
  MRI $+$ & -1      & 21.64 &      21.20 &      21.83 &      22.75 &      26.08 &       58.07 &       43.59 &       36.69 &       32.17 &       27.19 \\
  COGN   & (Annu.)  & (0.17) &    (0.19) &    (0.21) &    (0.22) &    (0.25) &      (0.1) &      (0.1) &     (0.13) &     (0.18) &     (0.19) \\
  MRI $+$ & -2      & 21.93 &      21.42 &      22.09 &      23.09 &      26.16 &       57.98 &       43.53 &       36.70 &       32.28 &       27.43 \\
  COGN   & (Annu.)  & (0.17) &    (0.19) &    (0.21) &    (0.23) &    (0.25) &      (0.1) &      (0.1) &     (0.14) &     (0.18) &     (0.19) \\
  MRI $+$ & -3      & 22.00 &      21.45 &      22.05 &      22.95 &      26.01 &       57.97 &       43.49 &       36.79 &       32.33 &       27.49 \\
  COGN   & (Annu.)  & (0.17) &    (0.19) &    (0.21) &    (0.22) &    (0.24) &      (0.1) &      (0.1) &     (0.14) &     (0.18) &     (0.19) \\
  MRI $+$ & -2      & 21.70 &      21.40 &      21.82 &      23.09 &      25.79 &       58.10 &       43.61 &       36.59 &       31.93 &       27.03 \\
  COGN   & (Bien.)  & (0.17) &    (0.19) &     (0.2) &    (0.22) &    (0.24) &      (0.1) &      (0.1) &     (0.13) &     (0.18) &     (0.19) \\
  \hline
  \end{tabular}}
\end{table}

\subsection{AUROC scores for the input modality case with cognitive tests (COGN) as the only longitudinal modality}
\label{app:cogn}
Please refer to Table~\ref{tab:rocauc_cogn}.

\begin{table}[hbtp]

  {\caption{AUROC scores obtained with {\shortname} under various patient history scenarios for the input modality case with cognitive tests (COGN) as the only longitudinal modality. History scenario format is ``history start year (data collection frequency)", year 0 being the {\now}. AUROC format is ``mean $\pm$ standard error".}\label{tab:rocauc_cogn}}  
  {\begin{tabular}{l|l|l|l|l|l|l|l|l|l|l|l}
  \hline
  Long. & \multicolumn{1}{l|}{History} & \multicolumn{5}{l|}{\bfseries CN-to-MCI conversion} & \multicolumn{5}{l}{\bfseries MCI-to-AD conversion }\\\cline{3-12}
  mod. & scenario & 1-year & 2-year & 3-year & 4-year & 5-year & 1-year & 2-year & 3-year & 4-year & 5-year\\
  \hline
  COGN & 0       & 81.15 &        81.45 &        79.02 &        77.38 &        78.32 &         85.66 &         88.15 &         90.88 &         92.60 &         92.86 \\
  ~   & -        & (0.45) &      (0.39) &      (0.34) &      (0.37) &      (0.36) &       (0.16) &       (0.16) &       (0.15) &       (0.15) &       (0.16) \\
  \hline
  COGN & -1      & 82.23 &        81.72 &        79.86 &        77.67 &        79.62 &         85.85 &         88.47 &         90.98 &         92.63 &         93.03 \\
  ~   & (Annu.)  & (0.44) &      (0.38) &      (0.34) &      (0.38) &      (0.37) &       (0.16) &       (0.16) &       (0.16) &       (0.16) &       (0.16) \\
  \hline
  COGN & -2      & 82.46 &        81.71 &        79.87 &        77.90 &        79.59 &         85.87 &         88.49 &         90.95 &         92.66 &         93.00 \\
  ~   & (Annu.)  & (0.44) &      (0.39) &      (0.35) &      (0.39) &      (0.38) &       (0.16) &       (0.16) &       (0.16) &       (0.15) &       (0.16) \\
  \hline
  COGN & -3      & 82.37 &        81.59 &        79.61 &        77.55 &        79.29 &         85.93 &         88.51 &         91.01 &         92.67 &         93.02 \\
  ~   & (Annu.)  & (0.44) &      (0.39) &      (0.35) &      (0.39) &      (0.38) &       (0.16) &       (0.16) &       (0.16) &       (0.16) &       (0.16) \\
  \hline
  COGN & -2      & 81.56 &        81.42 &        79.03 &        77.56 &        78.21 &         85.79 &         88.23 &         90.92 &         92.66 &         92.96 \\
  ~   & (Bien.)  & (0.44) &      (0.39) &      (0.34) &      (0.39) &      (0.38) &       (0.17) &       (0.16) &       (0.15) &       (0.15) &       (0.16) \\
  \hline
  \end{tabular}}
\end{table}

\subsection{Impact of dataset expansion during training}
\label{app:temp_train}
Table~\ref{tab:temporal_train} presents the AUROC scores obtained with {\shortname} under various patient history scenarios for the input modality case with MRI and cognitive tests (MRI$+$COGN) as the longitudinal modalities (complete-set case), without the dataset expansion method applied during training. 
Comparing with Table~\ref{tab:rocauc_comb}, we can see that the results in Table~\ref{tab:temporal_train} indicate signs of overfitting, as adding longer and more frequent longitudinal history data leads to a reduction in the model's predictive performance in every history scenario compared to the no-history scenario in MCI-to-AD conversion. 
Similarly, more frequent longitudinal data and longer history durations in various scenarios show significant performance reductions in the CN-to-MCI conversion. 
These findings underscore the importance of dataset expansion during training.

\begin{table}[hbtp]

  {\caption{AUROC scores obtained with {\shortname} under various patient history scenarios for the input modality case with MRI and cognitive tests (MRI$+$COGN) as the longitudinal modalities (complete-set case), without the dataset expansion method applied during training. History scenario format is ``history start year (data collection frequency)", year 0 being the {\now}. Specificity format is ``mean $\pm$ standard error".}\label{tab:temporal_train}}  
  {\begin{tabular}{l|l|l|l|l|l|l|l|l|l|l|l}
  \hline
  Long. & \multicolumn{1}{l|}{History} & \multicolumn{5}{l|}{\bfseries CN-to-MCI conversion} & \multicolumn{5}{l}{\bfseries MCI-to-AD conversion }\\\cline{3-12}
  mod. & scenario & 1-year & 2-year & 3-year & 4-year & 5-year & 1-year & 2-year & 3-year & 4-year & 5-year\\
  \hline
  MRI $+$ & 0       & 80.74 &        79.23 &        77.71 &        75.89 &        77.38 &         80.99 &         82.99 &         86.36 &         88.55 &         89.79 \\
  COGN   & -        & (0.82) &      (0.69) &      (0.63) &      (0.66) &      (0.74) &       (0.41) &       (0.39) &       (0.34) &       (0.36) &       (0.37) \\
  \hline
  MRI $+$ & -1      & 79.46 &        77.41 &        76.92 &        75.16 &        76.84 &         80.32 &         82.55 &         86.01 &         87.83 &         89.12 \\
  COGN   & (Annu.)  & (1.74) &      (1.27) &      (1.83) &      (1.41) &      (2.12) &       (0.84) &       (1.05) &       (0.97) &       (1.09) &       (1.29) \\
  \hline
  MRI $+$ & -2      & 80.04 &        78.09 &        77.16 &        75.33 &        77.47 &         80.20 &         81.97 &         85.25 &         87.10 &         88.01 \\
  COGN   & (Annu.)  & (2.1) &      (1.54) &      (2.08) &      (1.72) &      (2.27) &       (1.15) &       (1.32) &       (1.24) &       (1.36) &       (1.63) \\
  \hline
  MRI $+$ & -3      & 81.52 &        78.79 &        78.31 &        75.87 &        78.25 &         80.64 &         82.41 &         85.59 &         87.39 &         88.64 \\
  COGN   & (Annu.)  & (1.32) &       (1.0) &      (1.25) &      (1.06) &      (1.36) &        (0.7) &       (0.81) &       (0.76) &       (0.83) &       (0.99) \\
  \hline
  MRI $+$ & -2      & 81.43 &        79.67 &        78.56 &        77.06 &        77.89 &         80.81 &         83.03 &         86.14 &         88.34 &         89.42 \\
  COGN   & (Bien.)  & (0.88) &      (0.67) &      (0.65) &      (0.68) &      (0.85) &        (0.4) &       (0.39) &       (0.33) &       (0.39) &       (0.38) \\
  \hline
  \end{tabular}}
\end{table}

\subsection{Impact of temporal bias reduction during evaluation.}
\label{app:temp_test}
Table~\ref{tab:temporal_test} presents the AUROC scores obtained with {\shortname} under various patient history scenarios for the input modality case with MRI and cognitive tests (MRI$+$COGN) as the longitudinal modalities (complete-set case), evaluated without the temporal bias reduction.
Comparing the results in Table~\ref{tab:rocauc_comb} and Table~\ref{tab:temporal_test}, we observe significant variations in performance as fewer visits and, consequently, fewer disease stage samples are used in Table~\ref{tab:temporal_test}. This discrepancy highlights the presence of temporal bias in the data.

\begin{table}[hbtp]

  {\caption{AUROC scores obtained with {\shortname} under various patient history scenarios for the input modality case with MRI and cognitive tests (MRI$+$COGN) as the longitudinal modalities (complete-set case), evaluated without the temporal bias reduction. History scenario format is ``history start year (data collection frequency)", year 0 being the {\now}. Specificity format is ``mean $\pm$ standard error".}\label{tab:temporal_test}}  
  {\begin{tabular}{l|l|l|l|l|l|l|l|l|l|l|l}
  \hline
  Long. & \multicolumn{1}{l|}{History} & \multicolumn{5}{l|}{\bfseries CN-to-MCI conversion} & \multicolumn{5}{l}{\bfseries MCI-to-AD conversion }\\\cline{3-12}
  mod. & scenario & 1-year & 2-year & 3-year & 4-year & 5-year & 1-year & 2-year & 3-year & 4-year & 5-year\\
  \hline
  MRI $+$ & 0       & 84.72 &        84.21 &        85.07 &        74.03 &        74.58 &         90.70 &         87.35 &         84.67 &         86.87 &         84.32 \\
  COGN   & -        & (0.77) &      (1.36) &      (0.93) &      (1.03) &      (1.03) &       (0.38) &       (0.51) &       (0.48) &       (0.44) &       (0.48) \\
  \hline
  MRI $+$ & -1      & 90.64 &        85.35 &        84.49 &        75.40 &        74.91 &         90.67 &         88.34 &         85.25 &         86.33 &         84.50 \\
  COGN   & (Annu.)  & (0.61) &      (1.15) &       (1.0) &      (1.05) &      (1.01) &       (0.37) &       (0.48) &       (0.47) &       (0.46) &       (0.48) \\
  \hline
  MRI $+$ & -2      & 85.99 &        84.72 &        84.28 &        76.43 &        76.18 &         90.73 &         88.03 &         84.82 &         86.60 &         84.82 \\
  COGN   & (Annu.)  & (0.87) &      (0.98) &      (0.88) &       (1.0) &      (0.98) &       (0.38) &        (0.5) &       (0.49) &       (0.45) &       (0.47) \\
  \hline
  MRI $+$ & -3      & 84.72 &        84.36 &        84.29 &        73.26 &        74.17 &         90.77 &         87.85 &         84.34 &         85.71 &         83.70 \\
  COGN   & (Annu.)  & (0.88) &      (1.17) &      (1.03) &      (1.11) &      (1.03) &       (0.38) &       (0.49) &       (0.49) &       (0.46) &       (0.49) \\
  \hline
  MRI $+$ & -2      & 84.59 &        84.79 &        84.39 &        76.31 &        76.05 &         90.78 &         88.03 &         84.78 &         86.71 &         84.90 \\
  COGN   & (Bien.)  & (0.95) &      (0.97) &      (0.84) &       (1.0) &       (1.0) &       (0.38) &        (0.5) &       (0.49) &       (0.45) &       (0.47) \\
  \hline
  \end{tabular}}
\end{table}

%
%
%